\documentclass{article}
\usepackage{spconf,amsmath,graphicx,hyperref,amsfonts, amssymb}
\usepackage{adjustbox} 
\usepackage{diagbox}
\usepackage{pifont}
\usepackage{multirow}
\usepackage{threeparttable} 
\usepackage[table]{xcolor}
\usepackage[dvipsnames]{xcolor}
\usepackage{pgf}          
\usepackage{collcell}     

\newcommand{\cellcolorval}[1]{%
  \pgfmathsetmacro{\val}{1-#1}
  \ifdim\val pt<.5pt
    \pgfmathsetmacro{\g}{2*\val}%
    \pgfmathsetmacro{\r}{1}%
    \pgfmathsetmacro{\b}{0}%
  \else
    \pgfmathsetmacro{\t}{2*(\val-0.5)}%
    \pgfmathsetmacro{\r}{1-\t}%
    \pgfmathsetmacro{\g}{1}%
    \pgfmathsetmacro{\b}{0}%
  \fi
  \pgfmathsetmacro{\lum}{0.299*\r+0.587*\g+0.114*\b}%
  \ifdim\lum pt<0.5pt \def\textcol{white}\else\def\textcol{black}\fi
  \edef\colorcmd{\noexpand\cellcolor[rgb]{\r,\g,\b}}%
  \pgfmathsetmacro{\val}{#1}
  \colorcmd\textcolor{\textcol}{\pgfmathprintnumber[fixed,precision=2]{\val}}%
}

\title{Diagonal Artifacts in Samsung Images: PRNU Challenges and Solutions}
\name{David Vázquez-Padín, Fernando Pérez-González, Alejandro Martín-Del-Río\thanks{This work is supported in part by Xunta de Galicia and the European Regional Development Fund under Project ED431C 2025/41.}}
\address{atlanTTic Research Center, University of Vigo, E.E. de Telecomunicación, 36310 Vigo, Spain}
\begin{document}
\ninept
\maketitle
\begin{abstract}
We investigate diagonal artifacts present in images captured by several Samsung smartphones and their impact on PRNU-based camera source verification. We first show that certain Galaxy S series models share a common pattern causing fingerprint collisions, with a similar issue also found in some Galaxy A models. Next, we demonstrate that reliable PRNU verification remains feasible for devices supporting PRO mode with raw capture, since raw images bypass the processing pipeline that introduces artifacts. This option, however, is not available for the mid-range A series models or in forensic cases without access to raw images. Finally, we outline potential forensic applications of the diagonal artifacts, such as reducing misdetections in HDR images and localizing regions affected by synthetic bokeh in portrait-mode images.
\end{abstract}
\section{Introduction}
\label{sec:intro}

In image forensics, the de facto standard for verifying whether an image originates from a given camera is the sensor pattern noise known as Photo-Response Non-Uniformity (PRNU). Although PRNU has been successfully applied in forensic investigations, its reliability is increasingly challenged by advances in computational photography. Early inconsistencies stemmed from in-camera operations such as distortion correction \cite{MONTIBELLER_2024} and High Dynamic Range (HDR) processing \cite{MORTEZA_2019}, which required specialized compensation techniques. More recently, a critical issue has emerged in modern smartphones: the loss of PRNU uniqueness \cite{IULIANI_2021}. This results from sophisticated computational photography and AI-based enhancements, which introduce Non-Unique Artifacts (NUAs) that can cause false positive attributions. These phenomena undermine the assumption of the PRNU as a unique identifier and raise significant concerns about its continued role as the forensic standard \cite{KLIER_2025}.

Following the work of Iuliani \emph{et al.} \cite{IULIANI_2021}, which questioned the long-assumed uniqueness of PRNU, several studies have investigated the origins of the reported collisions. For Digital Single-Lens Reflex (DSLR) cameras, Butora and Bas \cite{BUTORA_2024} showed that many false positives arise from software-specific patterns introduced by Adobe during raw image development. For smartphones, Baracchi \emph{et al.} \cite{BARACCHI_2021} initiated the study of Non-Unique Artifacts (NUAs), focusing on the synthetic pattern embedded in Apple portrait-mode images, which we further characterized in \cite{VAZQUEZPADIN_2025}. Almost in parallel, Albisani \emph{et al.} \cite{ALBISANI_2021} identified NUAs in Samsung, Huawei, and Apple devices, while Bhat and Bianchi \cite{BHAT_2022} centered their analysis on Samsung, Huawei, and Xiaomi smartphones. These approaches enable practitioners to flag NUA-affected images by either excluding them from reference fingerprint estimation or lowering confidence in their attribution.

While identifying NUAs is an essential step to avoid collisions, Albisani \emph{et al.} \cite{ALBISANI_2021} reported that 100\% of the analyzed blocks from flat-field images captured by different Samsung models (i.e., S9 and S9+) were correlated, preventing the extraction of any valid fingerprint. To address this limitation, we first distinguish between the different sources of artifacts that explain the unexpectedly high correlations observed in \cite{ALBISANI_2021} versus those reported in \cite{BHAT_2022}, and then propose a solution to circumvent the issue highlighted in \cite{ALBISANI_2021}. More concretely, our main contributions are as follows:
\begin{itemize}
  \item 
  We uncover diagonal correlation patterns in various Samsung smartphones (including multiple Galaxy S and A models), which induce false positives in PRNU-based source camera verification.
  \item We demonstrate that, despite the widespread presence of such patterns, reliable PRNU-based verification remains feasible on devices supporting PRO mode with raw capture (typically high-end models), as raw images bypass the processing pipeline that introduces these artifacts.
  \item We highlight potential forensic uses of the identified diagonal patterns, such as mitigating misdetections in HDR images and localizing regions affected by the synthetic bokeh effect in portrait images.
\end{itemize}

The paper is organized as follows. Sect.~\ref{sec:preliminaries} briefly reviews PRNU-based camera source verification, followed by Sect.~\ref{sec:samsung_image_patterns}, which describes diagonal artifacts observed in Samsung devices. Sect.~\ref{sec:undoing_fingerprint_collisions} presents experiments showing that raw images prevent false positives and that the artifacts' diagonal correlations help mitigate HDR misdetections and localize bokeh regions. Finally, Sect.~\ref{sec:conclusions} concludes and outlines future work.

\section{PRNU-based Camera Source Verification}
\label{sec:preliminaries}

Luk{\'a}š \emph{et al.}~\cite{LUKAS_2005} introduced the use of the PRNU, a sensor-specific noise pattern, for verifying whether a given image was captured by a specific camera. The assumed model of the sensor output for an $H\times W$ single-channel image $\mathbf{Y}$ is $\mathbf{Y} = (\mathbf{1}+\mathbf{K})\circ \mathbf{X} + \mathbf{\Theta}$, where $\mathbf{K}$ is the PRNU, $\mathbf{X}$ the scene intensity, and $\mathbf{\Theta}$ other noise sources~\cite{CHEN_2008}. Based on this model, an estimate of a camera's PRNU fingerprint $\hat{\mathbf{K}}$ can be obtained from $L$ images $\{\mathbf{Y}_l\}_{l=1}^{L}$ by first denoising each image, i.e., $F(\mathbf{Y}_l)$ (where the filtering $F(\cdot)$ comes from \cite{MIHCAK_1999}), and computing residues $\mathbf{W}_l=\mathbf{Y}_l-F(\mathbf{Y}_l)$. The PRNU is then estimated via~\cite{CHEN_2008}:
\begin{equation}
    \hat{\mathbf{K}}=\left(\sum_{l=1}^{L}\mathbf{W}_l\circ\mathbf{Y}_l\right)\circ\left(\sum_{l=1}^{L}\mathbf{Y}_l\circ\mathbf{Y}_l\right)^{\circ-1},
    \label{eq:hat_K}
\end{equation}
where $\circ$ denotes the element-wise (Hadamard) product and $(\cdot)^{\circ-1}$ the element-wise inverse.

Now, given a test image $\mathbf{Y}$ with residue $\mathbf{W}=\mathbf{Y}-F(\mathbf{Y})$ and a PRNU estimate $\hat{\mathbf{K}}$, camera attribution can be framed as a binary hypothesis test. The null hypothesis, $\mathcal{H}_0$, assumes that $\mathbf{W}$ does not contain the PRNU $\mathbf{K}$, while the alternative, $\mathcal{H}_1$, assumes that it does. If $\mathcal{H}_1$ holds, the test image $\mathbf{Y}$ likely originates from the camera with PRNU $\mathbf{K}$. In practice, the hypothesis test is evaluated using the Peak-to-Correlation Energy (PCE) \cite{GOLJAN_2009}, which is defined as follows 
\begin{equation}
  \text{PCE}(\mathbf{W},\hat{\mathbf{K}}\circ\mathbf{Y})\triangleq \frac{(HW-|\mathcal{N}|)\cdot\text{sgn}(\rho_{\max})\cdot(\rho_{\max})^2}{\sum_{(s_1,s_2)\notin\mathcal{N}}(\rho_{\mathbf{W},\hat{\mathbf{K}}\circ\mathbf{Y}}(s_1,s_2))^2},
  \label{eq:PCE}
\end{equation}
where $\rho_{\mathbf{A},\mathbf{B}}(s_1,s_2)$ denotes the Normalized Cross-Correlation (NCC) between $\mathbf{A}$ (with sample mean $\bar{\mathbf{A}}$) and $\mathbf{B}$ (with sample mean $\bar{\mathbf{B}}$) for a given shift $(s_1,s_2)$, i.e.,
\begin{equation*}
  \rho_{\mathbf{A},\mathbf{B}}(s_1,s_2)\triangleq\frac{\sum_{i,j}(A_{i,j}\hspace{-0.15em}-\hspace{-0.15em}\bar{\mathbf{A}})\cdot(B_{i+s_i,j+s_2}\hspace{-0.15em}-\hspace{-0.15em}\bar{\mathbf{B}})}{(\sum_{i,j}(A_{i,j}\hspace{-0.15em}-\hspace{-0.15em}\bar{\mathbf{A}})^2\sum_{i,j}(B_{i+s_i,j+s_2}\hspace{-0.15em}-\hspace{-0.15em}\bar{\mathbf{B}})^2)^{\frac{1}{2}}},
\end{equation*}
and $\rho_{\max}\triangleq\max_{(s_1,s_2)\in\mathcal{S}}\rho_{\mathbf{W},\hat{\mathbf{K}}\circ\mathbf{Y}}(s_1,s_2)$ with $\mathcal{S}$ denoting all possible pixel coordinates in the test image. In \eqref{eq:PCE}, $\mathcal{N}$ denotes a small neighborhood around the peak NCC of size $11\times11$. Finally, by evaluating $\text{PCE}(\mathbf{W},\hat{\mathbf{K}}\circ\mathbf{Y})\underset{\mathcal{H}_0}{\overset{\mathcal{H}_1}{\gtrless}}\tau$ for a given threshold $\tau$, the decision is made. A threshold of 60 is typically adopted, as validated in~\cite{GOLJAN_2009} on over a million images from 6,896 cameras, achieving a false alarm rate of $10^{-5}$ and a detection rate of $97.6\%$.

\section{Samsung Image Artifacts}
\label{sec:samsung_image_patterns}
Previous works by Albisani \emph{et al.} \cite{ALBISANI_2021} and Bhat and Bianchi \cite{BHAT_2022} showed the presence of NUAs in images captured by Samsung devices (mainly from the Galaxy S series). 
More recently, Montibeller \emph{et al.} \cite{MONTIBELLER_2024b} reported that spikes in the Fourier domain appear to be correlated with the absence of NUAs responsible for false positives in Samsung A50 devices. Although the specific artifacts identified in previous studies differ, our analysis of mismatches across Samsung Galaxy S and A5* devices shows that Samsung embeds distinct, model-dependent artifacts of unclear origin that exhibit stronger correlations than the PRNU.

\begin{table}[t]
  \caption{List of selected Samsung devices.}
  \label{tab:list_selected_devices}
  \centering
  \begin{adjustbox}{width=\linewidth}
    \begin{threeparttable}
      \begin{tabular}{c|c|c|c|c|c|c}
        \hline\hline
        \multirow{2}{*}{\textbf{ID}} & \multirow{2}{*}{\textbf{Model}} & \multirow{2}{*}{\textbf{Version}} & \multicolumn{2}{c|}{\textbf{\#images}} & \textbf{Resolution} & \multirow{2}{*}{\textbf{Source}}\\
        & & & \textbf{ref.} & \textbf{test} & $(H\times W)$&\\
        \hline
        \multicolumn{7}{c}{USA Version (Galaxy S Series)}\\
        \hline
        \textemdash & S9 & SM-G960U & \textemdash & 245 & 3024$\times$4032 & \cite{IULIANI_2021}\\
        \textemdash & S9+ & SM-G965U & \textemdash & 257 & 3024$\times$4032 & \cite{IULIANI_2021}\\
        \textemdash & S10 & SM-G973U & \textemdash & 133 & 3024$\times$4032 & \cite{IULIANI_2021}\\
        \textemdash & S10+ & SM-G975U & \textemdash & 228 & 3024$\times$4032 & \cite{IULIANI_2021}\\
        \hline
        \multicolumn{7}{c}{Global Version (Galaxy S Series)}\\
        \hline
        SD01 & S9 & SM-G960F & 5 & 30 & 3024$\times$4032 & C14 \cite{ALBISANI_2021}\\
        SD02 & S9 & SM-G960F & \textemdash & 9 & 3024$\times$4032 & \cite{GSMARENA}\\
        SD03 & S9+ & SM-G965F & 5 & 40 & 3024$\times$4032 & C15 \cite{ALBISANI_2021}\\
        SD04 & S9+ & SM-G965F & \textemdash & 8 & 3024$\times$4032 & \cite{GSMARENA}\\
        SD05 & S10 & SM-G973F & 54 & 108 & 3024$\times$4032 & D44 \cite{FLOREVIEW_2023}\\
        SD06 & S10 & SM-G973F & 40$^\dagger$ & 112$^*$ & 3024$\times$4032 & Ours\\
        SD07 & S10 & SM-G973F & \textemdash & 16 & 3024$\times$4032 & \cite{GSMARENA}\\
        SD08 & S10+ & SM-G975F & 32 & 108 & 3024$\times$4032 & D30 \cite{FLOREVIEW_2023}\\
        \hline
        \multicolumn{7}{c}{Galayx A5* Series}\\
        \hline
        AD01 & A53 & SM-A536B & 12 & 14 & 3468$\times$4624 & \cite{FLICKR_API}\\
        AD02 & A53 & SM-A536B & \textemdash & 9 & 3468$\times$4624 & \cite{GSMARENA}\\
        AD03 & A54 & SM-A546B & 9 & 17 & 3060$\times$4080 & Ours\\
        AD04 & A54 & SM-A546B & \textemdash & 18 & 3060$\times$4080 & \cite{GSMARENA}\\
        AD05 & A55 & SM-A556B & 22 & 19 & 3060$\times$4080 & \cite{FLICKR_API}\\
        AD06 & A55 & SM-A556B & \textemdash & 25 & 3060$\times$4080 & \cite{GSMARENA}\\
        AD07 & A56 & SM-A566B & 20 & 62 & 3000$\times$4000 & Ours\\
        AD08 & A56 & SM-A566B & \textemdash & 13 & 3000$\times$4000 & \cite{GSMARENA}\\
        \hline\hline
      \end{tabular}
      \begin{tablenotes}[para,flushleft]
        \scriptsize
        \item $^\dagger$ 20 (out of 40) are captured under the PRO mode.
        \item $^*$ 41 (out of 112) are teken with the telephoto sensor.
      \end{tablenotes}
    \end{threeparttable}
  \end{adjustbox}
\end{table}

\begin{figure}[t]
  \begin{minipage}[b]{.49\linewidth}
    \centering
    \centerline{\includegraphics[width=\linewidth]{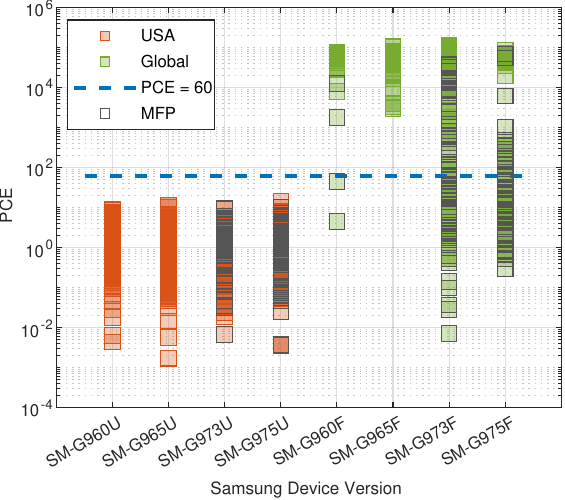}}
    \centerline{\footnotesize (a) Galaxy S series}
  \end{minipage}
  \begin{minipage}[b]{.49\linewidth}
    \centering
    \centerline{\includegraphics[width=\linewidth]{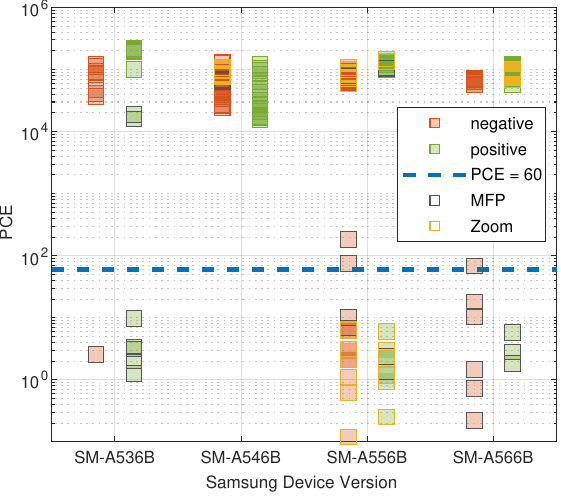}}
    \centerline{\footnotesize (b) Galaxy A5* series}
  \end{minipage}
  \caption{Scatter plots of the PCE values obtained for the Galaxy S series devices (a) and A5* series (b).}
  \label{fig:PCE_values}
\end{figure}

In the following, we analyze a set of Samsung devices listed in Tab.~\ref{tab:list_selected_devices}. A subset of images is drawn from the dataset in \cite{IULIANI_2021}, where we aggregate devices by model and version without assigning individual IDs. Additional samples are collected from public datasets such as \cite{ALBISANI_2021} and FloreView \cite{FLOREVIEW_2023}, from out-of-the-camera smartphone pictures available in GSMarena reviews \cite{GSMARENA}, as well as from our own devices, which are explicitly labeled.\footnote{All data will be made publicly available upon acceptance of this paper.} For the Galaxy S series, we explicitly distinguish between the USA version (codenames ending in ``U'') and the Global version (codenames ending in ``F''), as they incorporate different chipsets (Qualcomm Snapdragon and Exynos, respectively) which likely lead to different imaging pipelines and, consequently, distinct forensic traces. Indeed, we observed that all Global version devices embed a specific diagonal artifact in images captured in default photo mode, which is distinct from their USA counterparts. To illustrate this, we extracted the PRNU fingerprint from our S10 device (SD06 in Tab.\ref{tab:list_selected_devices}), following the procedure described in Sect.~\ref{sec:preliminaries}, using 20 flat-field images (captured under the default photo mode) in \eqref{eq:hat_K}. We then computed the PCE values, as defined in \eqref{eq:PCE}, between this fingerprint and the residuals of all S series test images. The results, grouped by device version, are reported in the scatter plot of Fig.~\ref{fig:PCE_values}(a).

This scatter plot highlights two main findings. First, the pattern embedded by Global version models is uncorrelated with all USA version models. This is important because the studies in \cite{IULIANI_2021} and \cite{BHAT_2022} analyzed only USA models, so their conclusions apply exclusively to that version.\footnote{In this work, we do not further investigate NUAs in USA models, leaving such analysis to future research.} Second, within the Global version it is not possible to distinguish between specific devices or models, as most PCE values exceed the commonly used threshold of 60 (consistent with prior observations for the S9 (C14) and S9+ (C15) in \cite{ALBISANI_2021} and later for the S10 (D44) and S10+ (D30) in \cite{FLOREVIEW_2023}). Moreover, all PCE values from Global version devices that fall below this threshold correspond to images processed through a Multi-Frame Processing (MFP) pipeline, e.g., HDR rendering from stacked frames, and are labeled accordingly in Fig.~\ref{fig:PCE_values}(a).\footnote{MFP labeling was performed by searching the strings \texttt{MHDR}, \texttt{LHDR}, and \texttt{MFP3} in the JPEG APP4 marker that contains proprietary Samsung metadata.} Further details on HDR-related misdetections are provided in Sect.~\ref{subsec:analysis_of_HDR}.

\begin{figure}[t]
  \begin{minipage}[b]{.32\linewidth}
    \centering
    \centerline{\includegraphics[width=\linewidth]{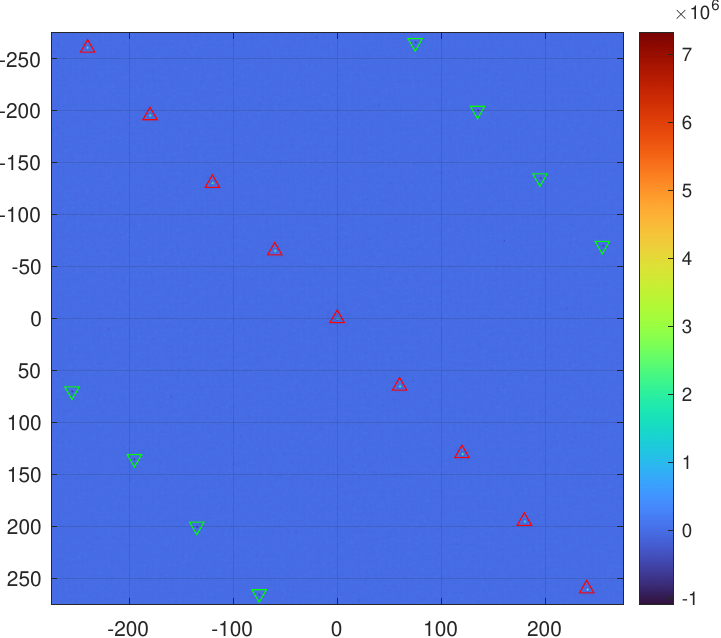}}
    \centerline{\footnotesize (a) S10 [SM-G973F]}
  \end{minipage}
  \begin{minipage}[b]{.32\linewidth}
    \centering
    \centerline{\includegraphics[width=\linewidth]{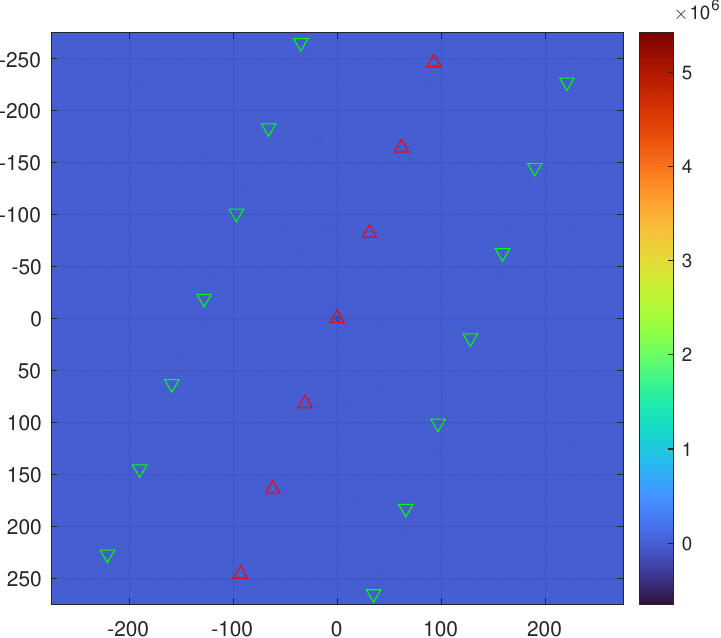}}
    \centerline{\footnotesize (b) A53 [SM-A536B]}
  \end{minipage}
  \begin{minipage}[b]{.32\linewidth}
    \centering
    \centerline{\includegraphics[width=\linewidth]{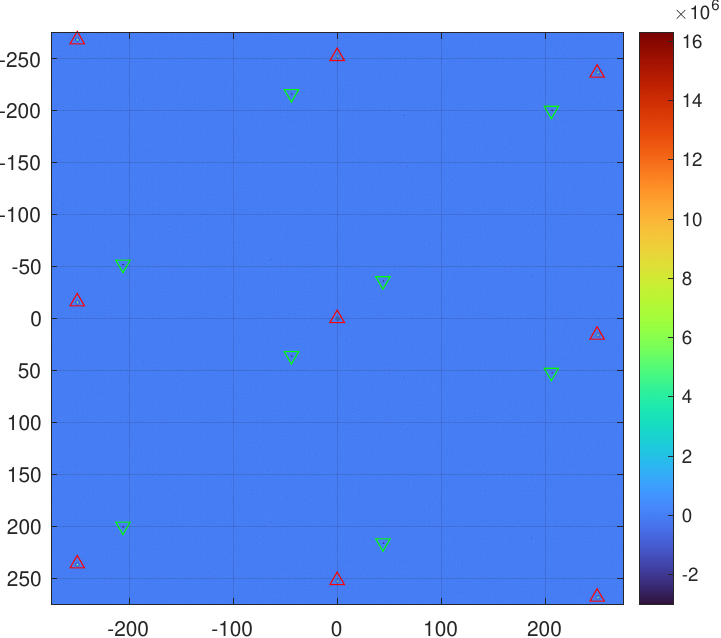}}
    \centerline{\footnotesize (c) A56 [SM-A566B]}
  \end{minipage}
  \caption{Autocorrelation of the estimated PRNUs from different devices: SD06 (a), AD01 (b), and AD07 (c). For clarity, visualization is limited to the central block of size $551\times551$.}
  \label{fig:samsung_patterns_autocorrelation}
\end{figure}

For Global version Galaxy S series devices SD01-SD08 listed in Tab.~\ref{tab:list_selected_devices}, all images with PCE values well above the threshold share a distinct feature: their autocorrelation reveals a diagonal periodicity that aligns with the autocorrelation of the PRNU fingerprint extracted from the SD06 device shown in Fig.~\ref{fig:samsung_patterns_autocorrelation}(a). To emphasize this diagonal structure, we restrict the visualization to the central block of size $551\times551$ and mark the strongest positive peaks with \textcolor{red}{\raisebox{1.3ex}{\rotatebox{180}{$\triangledown$}}} and the strongest negative peaks with \textcolor{green}{$\triangledown$}. Within this region, positive peaks appear, for instance, at integer multiples of the coordinates $[60,65]$, which suggests the involvement of block-based processing, although the exact origin of the periodicity remains unclear. 

Notice that Albisani \emph{et al.} \cite[Fig.~4]{ALBISANI_2021} previously reported the presence of a diagonal correlation pattern in the devices S9 and S9+. In our analysis, we further observe that periodic negative correlations are also present, and that these artifacts are not limited to the Global version S9 and S9+ devices but extend also to the S10 and S10+. Moreover, in these later devices equipped with three sensors (wide, telephoto, and ultrawide), the same correlation pattern consistently emerges in images captured with the wide and telephoto sensors (which have different PRNUs) when captured in default photo mode.\footnote{The ultrawide sensor is excluded from this analysis because its separate processing pipeline, including distortion correction and a different resolution ($3456\times4608$ with respect to $3024\times4032$), produces a distinct diagonal pattern, which will be investigated in future work.} Regarding the spatial localization of these artifacts within an image, we found that correlation values are higher in flat regions and weaker in textured areas, likely due to denoising filters. Specific examples illustrating these cases are provided in the technical report \cite[Sect.~2]{TECHREP}.

Focusing now on Galaxy A5* series devices AD01-AD08 listed in Tab.\ref{tab:list_selected_devices}, we analyzed default photo mode images from two devices per model. PRNU fingerprints were extracted as in Sect.~\ref{sec:preliminaries} using reference images from AD01, AD03, AD05, and AD07. The resulting PCE values, grouped by device version, are shown in the scatter plot of Fig.~\ref{fig:PCE_values}(b), with positives in green and negatives in red. Regardless of the class (positive/negative), most values cluster above $10^4$, indicating the presence of common artifacts similar to that of the S series. As before, we use different marker border colors to distinguish samples obtained from MFP images (e.g., HDR) and digitally zoomed images (determined from EXIF metadata), as these can produce PCE values at or below the threshold if not corrected. The extracted fingerprints again show diagonal structures in their autocorrelation, illustrated for the A53 (AD01) in Fig.~\ref{fig:samsung_patterns_autocorrelation}(b) and the A56 (AD07) in Fig.~\ref{fig:samsung_patterns_autocorrelation}(c). Plots for A54 and A55 are omitted, as their fingerprints match A56's when resized from $4080\times3060$ to $4000\times3000$.

We extended the search for diagonal patterns to older S models (S8, S8+) and newer ones (S20–S24), as well as older A5* models (A50, A51). No matches were found, except for the mismatches reported in the FloreView dataset \cite[Tab.~5]{FLOREVIEW_2023} between an S10 and an S21+ (both Global versions). Using the fingerprint from our S10 (SD06), five S21+ images (D17 in \cite{FLOREVIEW_2023}) exceeded the PCE threshold of 60, yielding false positives. As noted in the technical report \cite[Sect.~3]{TECHREP}, local correlation analysis showed residual diagonal traces in flat regions (e.g., sky patches). While newer S models seem to have removed these artifacts, they persist in the A5* series, with the A56 (released in 2025) still exhibiting them. A summary of matches between diagonal artifacts across Samsung models is given in \cite[Tab.~1]{TECHREP}.

\section{Addressing S Series fingerprint collisions}
\label{sec:undoing_fingerprint_collisions}

So far, we have shown that capturing reference images in the default photo mode on any Global version S device listed in Tab.~\ref{tab:list_selected_devices} causes fingerprint collisions, making PRNU-based camera source verification ineffective. This raises the question of whether the PRNU is still present in these fingerprints, albeit at lower power. Experiments with our SD06 (S10) device show that saving images in HEIF (instead of JPEG) preserves the diagonal pattern, failing to prevent collisions. In contrast, capturing reference images in PRO mode produces raw images (in DNG format) without diagonal artifacts, yielding fingerprints that closely align with the sensor's PRNU.\footnote{Raw images in DNG format were converted to uncompressed RGB images using \texttt{dcraw}~\cite{DCRAW} with default settings.}

\begin{figure}[t]
  \begin{minipage}[c]{.49\linewidth}
    \centering
    \includegraphics[width=0.9\linewidth]{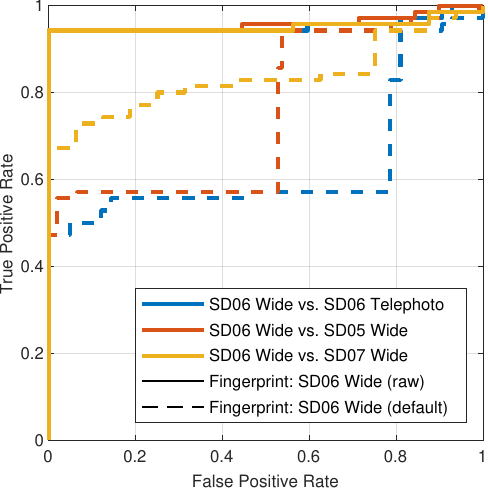}
    \centerline{\footnotesize (a)}
  \end{minipage}
  \begin{minipage}[c]{.49\linewidth}
    \begin{adjustbox}{width=\linewidth}
      \begin{threeparttable}
        \begin{tabular}{l|c|c}
          \hline\hline
          & \multicolumn{2}{c}{\textbf{FPR} values$^\dagger$}\\
          \cline{2-3}
          & $\begin{tabular}{@{}c@{}}\textbf{SD06}\\(default)\end{tabular}$ & $\begin{tabular}{@{}c@{}}\textbf{SD06}\\(raw)\end{tabular}$ \\ \hline
          \textbf{S9} & \cellcolorval{0.9487} & \cellcolorval{0.0}\\
          \textbf{S9+} & \cellcolorval{1.000} & \cellcolorval{0.0}\\
          \textbf{S10} & \cellcolorval{0.6855} & \cellcolorval{0.0}\\
          \textbf{S10+} & \cellcolorval{0.657} & \cellcolorval{0.0}\\
          \hline\hline
        \end{tabular}
        \begin{tablenotes}[para,flushleft]
          \item $^\dagger$ \textbf{TPR} = 0.94 in both cases
        \end{tablenotes}
      \end{threeparttable}
    \end{adjustbox}
    \vspace{0.06cm}

    \centerline{\footnotesize (b)}
  \end{minipage}
  \caption{ROC curves comparing performance with raw and default fingerprints (a). TPR and FPR values at $\tau=60$ (b).}
  \label{fig:ROC_results}
\end{figure}

Building on this finding, we tested whether the fingerprint extracted from PRO-mode raw images of the SD06 wide sensor could correctly identify default photo mode images from the same wide sensor (bearing the same PRNU) while rejecting default photo mode images from the SD06 telephoto sensor (same device and resolution, but different PRNU) and from the wide sensors of other S10 devices (SD05 and SD07). Fig.~\ref{fig:ROC_results}(a) shows the obtained ROC curves: the SD06 default fingerprint containing diagonal artifacts (dashed lines) performs poorly, as expected, while the PRO-mode raw fingerprint (solid lines) markedly improves distinguishability across sensors, though perfect detection is limited by HDR images. For the remaining S series models with known fingerprint collisions, the table in Fig.~\ref{fig:ROC_results}(b) shows that using the raw fingerprint yields a False Positive Rate (FPR) of 0 at the recommended PCE threshold of 60, while the True Positive Rate (TPR) for SD06 stays at 0.94. Perfect detection would require handling HDR images, as discussed below in Sect.~\ref{subsec:analysis_of_HDR}.


Unfortunately, this approach to resolving fingerprint collisions in S series devices is not applicable in forensic cases without access to the original device (needed to capture raw images),\footnote{Regrettably, the telephoto sensor cannot be used under the PRO mode of our S10 device and thus its relative raw fingerprint cannot be extracted.} nor in A5* series devices, whose mid-range hardware only supports JPEG output with the embedded pattern and lacks PRO-mode DNG capture. This limitation calls for alternative strategies to suppress the diagonal pattern in extracted fingerprints while preserving the low-power PRNU signal, a direction we will explore in future work.

\subsection{Analysis of HDR mode}
\label{subsec:analysis_of_HDR}

\begin{figure}[t]
\begin{minipage}[b]{.48\linewidth}
    \centering
    \centerline{\includegraphics[width=\linewidth]{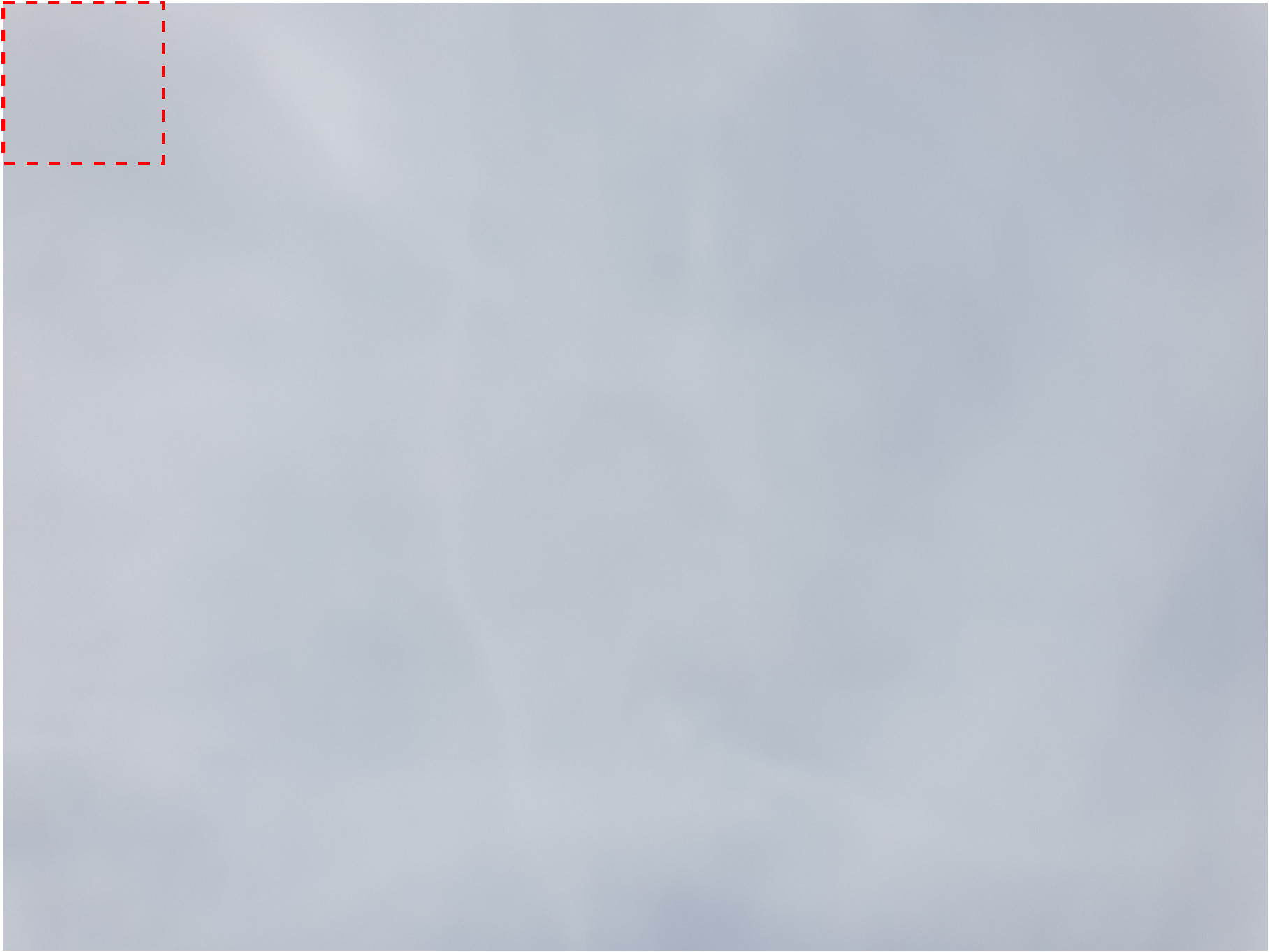}}
    \centerline{\footnotesize (a) Non-HDR image}
  \end{minipage}
  \begin{minipage}[b]{.48\linewidth}
    \centering
    \centerline{\includegraphics[width=\linewidth]{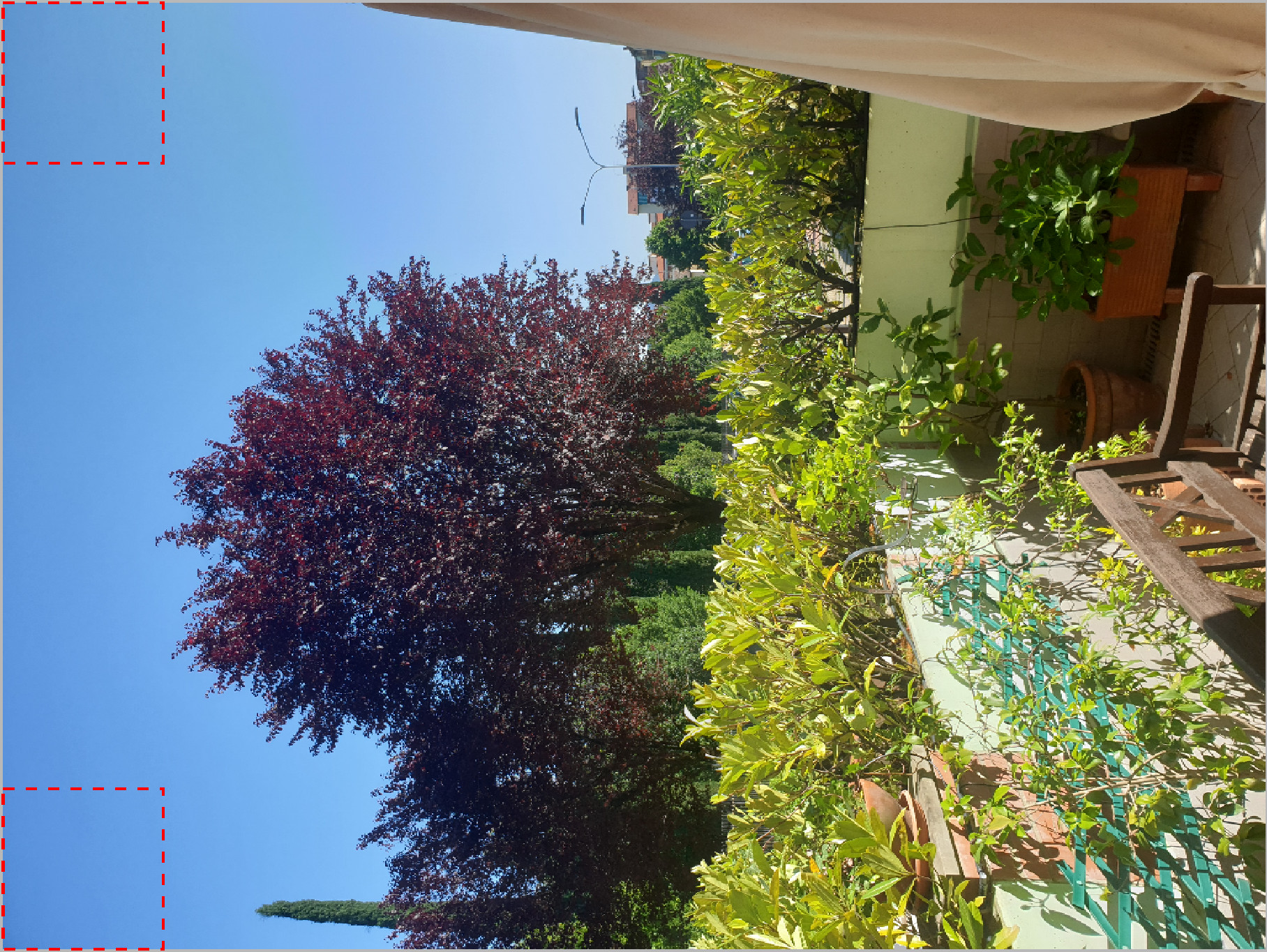}}
    \centerline{\footnotesize (b) HDR image}
  \end{minipage}
\\
  \begin{minipage}[b]{.325\linewidth}
    \centering
    \centerline{\includegraphics[width=\linewidth]{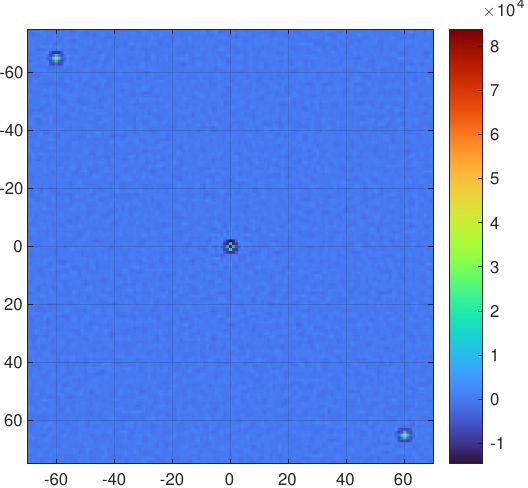}}
    \centerline{\footnotesize (c) Non-HDR (top left)}
  \end{minipage}
  \begin{minipage}[b]{.325\linewidth}
    \centering
    \centerline{\includegraphics[width=\linewidth]{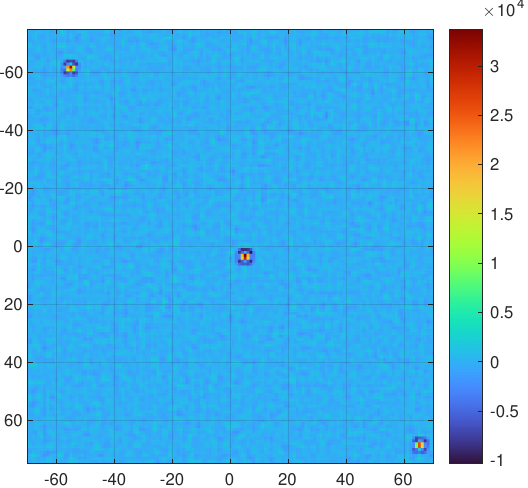}}
    \centerline{\footnotesize (d) HDR (top left)}
  \end{minipage}
  \begin{minipage}[b]{.325\linewidth}
    \centering
    \centerline{\includegraphics[width=\linewidth]{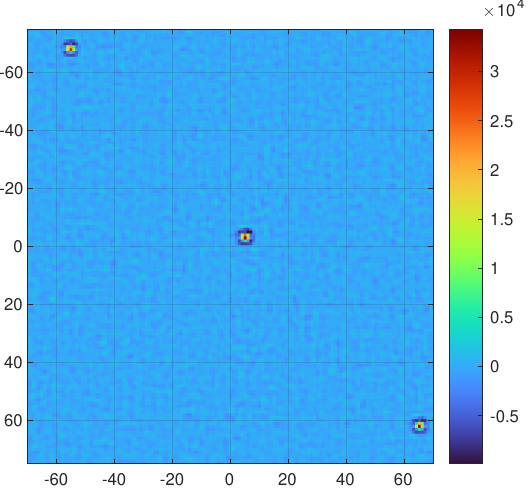}}
    \centerline{\footnotesize (e) HDR (bottom left)}
  \end{minipage}
  \caption{Examples of cross-correlation between the default fingerprint (with diagonal artifacts) and image residuals. Upper panel: non-HDR (a) and HDR (b) images from \cite{ALBISANI_2021}. Lower panel: non-HDR top-left block showing a peak at $[0,0]$ (c); HDR top-left block showing shifted peaks with maximum at $[5,3]$ (d); HDR bottom-left block with maximum at $[5,-3]$ (e), revealing local translations from HDR construction.}
  \label{fig:HDR}
\end{figure}

Although fingerprints containing diagonal artifacts are responsible for all reported collisions, they can still be leveraged positively to reveal localized translations introduced during HDR image construction \cite[Sect.~2]{MORTEZA_2019}, since the underlying diagonal correlation pattern shifts accordingly. To demonstrate this, we analyze the cross-correlation between blocks of the default fingerprint (with diagonal artifacts) and residuals from non-HDR and HDR images (examples in Fig.~\ref{fig:HDR}(a)-(b)). For the $512\times512$ top-left block of the non-HDR image (marked in red in Fig.~\ref{fig:HDR}(a)), Fig.~\ref{fig:HDR}(c) shows single peaks of the diagonal pattern, with the maximum at $[0,0]$. Similarly, for the top-left block in the HDR case (Fig.~\ref{fig:HDR}(b)), the cross-correlation (Fig.~\ref{fig:HDR}(d)) no longer shows a single peak; instead, nearby peaks of similar magnitude appear, with the maximum shifted to $[5,3]$. For the bottom-left HDR block (Fig.~\ref{fig:HDR}(e)), the maximum shifts to $[5,-3]$, indicating a local translation in that region. This suggests that a block-wise analysis across the entire frame could yield a map of local translations. Applying these translations to raw-extracted fingerprints would enable full-frame PCE computation against the adapted fingerprint, potentially reducing HDR misdetections (a strategy we plan to explore in future work).


\subsection{Analysis of Portrait mode}
\label{subsec:analysis_of_bokeh}

\begin{figure}[t]
  \begin{minipage}[b]{.24\linewidth}
    \centering
    \centerline{\includegraphics[height=2.355cm]{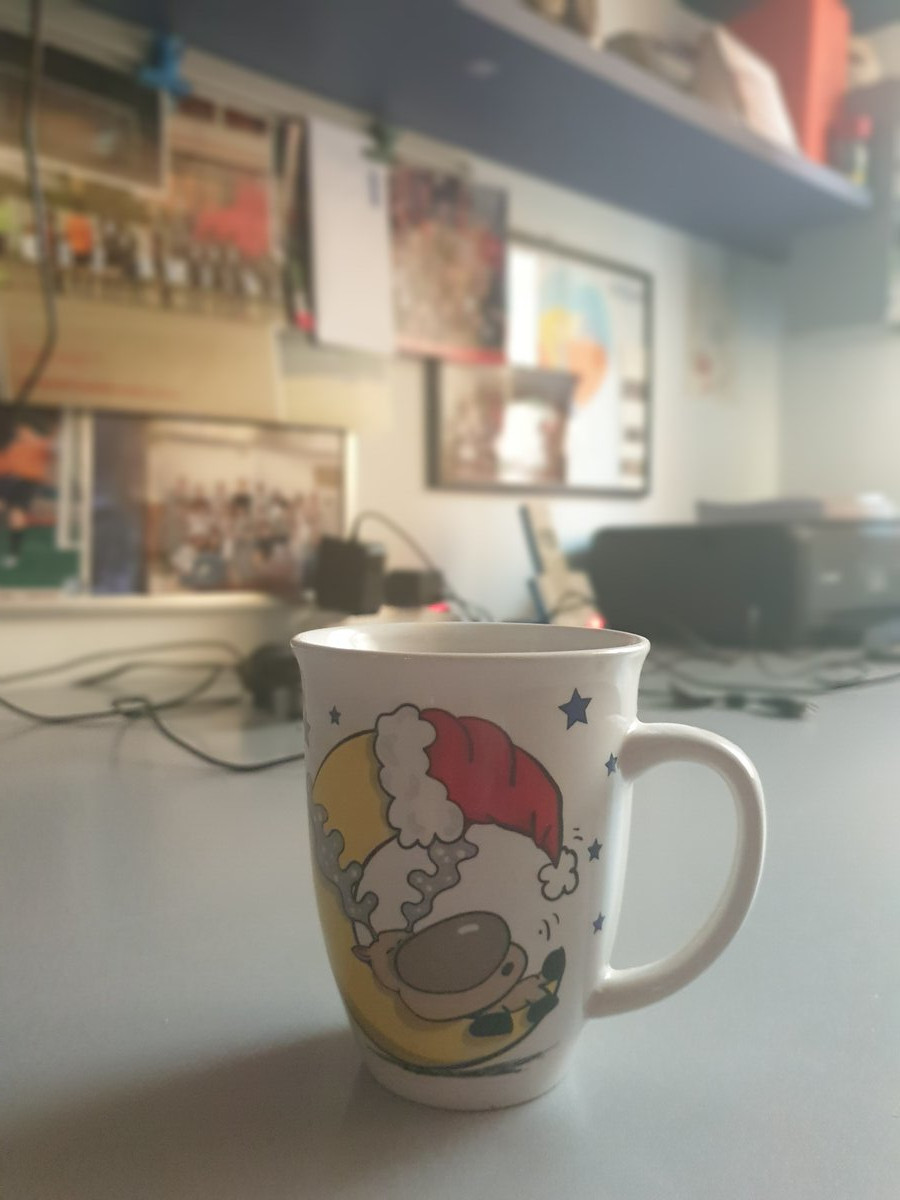}}
    \centerline{\footnotesize (a) Portrait image}
  \end{minipage}
  \begin{minipage}[b]{.24\linewidth}
    \centering
    \centerline{\includegraphics[height=2.355cm]{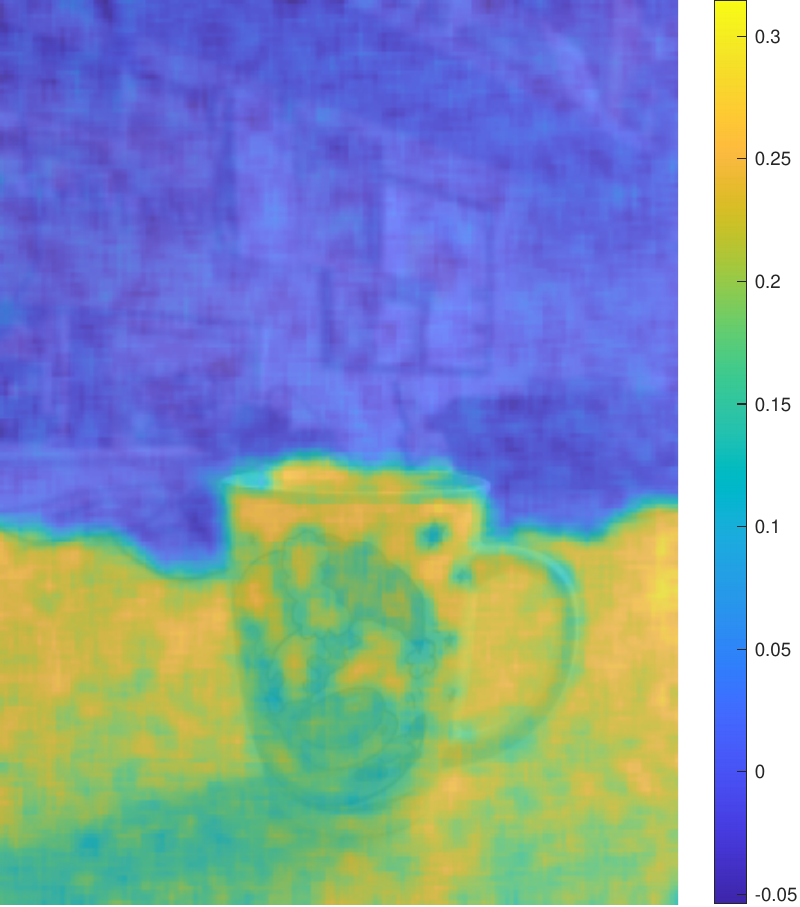}}
    \centerline{\footnotesize (b)}
  \end{minipage}
  \begin{minipage}[b]{.24\linewidth}
    \centering
    \centerline{\includegraphics[height=2.355cm]{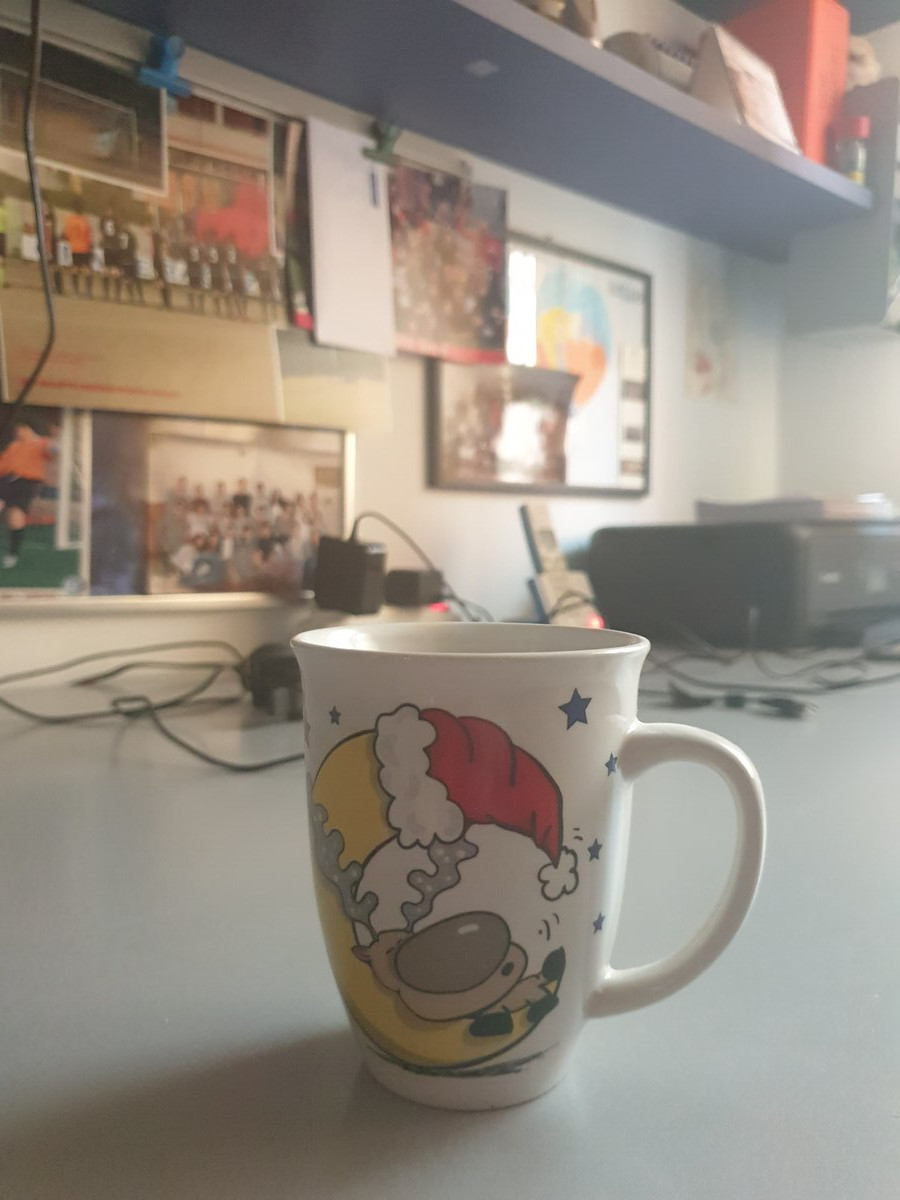}}
    \centerline{\footnotesize (c) Pre-bokeh image}
  \end{minipage}
  \begin{minipage}[b]{.24\linewidth}
    \centering
    \centerline{\includegraphics[height=2.355cm]{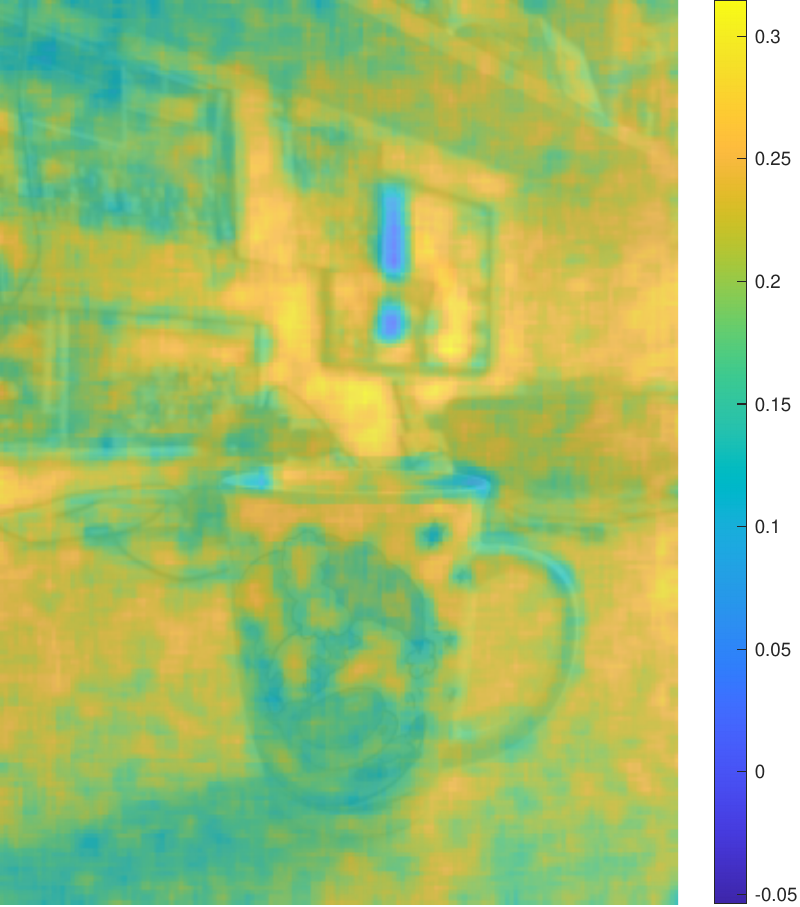}}
    \centerline{\footnotesize (d)}
  \end{minipage}
  \caption{Example of block-wise cross-correlation ($21\times21$ blocks) between the default fingerprint (with diagonal artifacts) and residuals from Samsung portrait images. Portrait image from \cite{ALBISANI_2021} with bokeh effect (a). Corresponding correlation map showing no correlation in bokeh regions (b). Original pre-bokeh image (c), exhibiting uniform correlation across the frame (d).}
  \label{fig:bokeh}
\end{figure}

Unlike Apple devices \cite{VAZQUEZPADIN_2025}, bokeh regions in Samsung portrait images do not correlate with the default fingerprint or across images. To demonstrate this, we computed a block-wise cross-correlation (block size $21\times21$) between the default fingerprint (with diagonal artifacts) and residuals from Samsung portrait images, such as the example in Fig.~\ref{fig:bokeh}(a). The resulting map (Fig.~\ref{fig:bokeh}(b)) shows no correlation in bokeh areas, whereas the original (pre-bokeh) image (Fig.~\ref{fig:bokeh}(c)) maintains relatively uniform correlation across the frame. This confirms the suspicion in \cite{ALBISANI_2021} that bokeh effects are unlikely to introduce false positives beyond those caused by the diagonal artifacts. However, they may still lead to misdetections with raw-extracted fingerprints, since the low-pass filtering and noise addition used to generate the bokeh effect degrade the underlying PRNU. Therefore, excluding bokeh regions during fingerprint extraction and detection is recommended, especially when only portrait images are available. A method that exploits the diagonal pattern to handle bokeh regions will be explored in future work.

\section{Conclusions}
\label{sec:conclusions}

In this work, we have identified artifacts embedded in images captured by various Samsung smartphone models, characterized by diagonal correlations. In the Galaxy S series, the Global version of the S9, S9+, S10, and S10+ devices share the same diagonal pattern, while in the S21+ model residual traces of the pattern are still locally observable. In contrast, the A5* series shows multiple patterns: the A53 presents a distinct diagonal pattern, while the A54, A55, and A56 (after resizing) share a common one. These patterns induce false positives (and thus fingerprint collisions, as reported in \cite{ALBISANI_2021}) when performing PRNU-based source camera verification. We expect that identifying these diagonal patterns and their relationships across distinct models will help clarify the origins of collisions observed in current and past forensic investigations.

Our key contribution is demonstrating that fingerprint collisions observed in the Global version of certain Galaxy S series models can be avoided. By using RAW images captured in PRO mode (which bypass the imaging pipeline responsible for introducing the diagonal correlations) we can extract a fingerprint that accurately reflects the sensor's PRNU and thus prevents collisions. However, this solution is not viable for the A5* series, which does not support raw capture in PRO mode, nor for forensic scenarios where raw images from the questioned device are unavailable. To address these limitations, future work will focus on developing methods to suppress the diagonal artifacts in extracted fingerprints while preserving the underlying low-power PRNU signal. In addition, we plan to extend our analysis to the US version of the S series models previously examined in \cite{BHAT_2022}.

Finally, we have outlined potential ways to exploit the diagonal pattern to address challenges in HDR and portrait mode images, which will be further developed in future research.
\bibliographystyle{IEEEbib}
\bibliography{samsung}

\begin{thebibliography}{10}

\bibitem{MONTIBELLER_2024}
Andrea Montibeller and Fernando Pérez-González,
\newblock ``{An Adaptive Method for Camera Attribution Under Complex Radial Distortion Corrections},''
\newblock {\em IEEE Transactions on Information Forensics and Security}, vol. 19, pp. 385--400, 2024.

\bibitem{MORTEZA_2019}
Morteza Darvish Morshedi~Hosseini and Miroslav Goljan,
\newblock ``{Camera Identification from HDR Images},''
\newblock in {\em Proceedings of the ACM Workshop on Information Hiding and Multimedia Security}, New York, NY, USA, 2019, IH\&MMSec'19, p. 69–76, Association for Computing Machinery.

\bibitem{IULIANI_2021}
Massimo Iuliani, Marco Fontani, and Alessandro Piva,
\newblock ``{A Leak in PRNU Based Source Identification. Questioning Fingerprint Uniqueness},''
\newblock {\em IEEE Access}, vol. 9, pp. 52455--52463, 2021.

\bibitem{KLIER_2025}
Samantha Klier and Harald Baier,
\newblock ``{Source Camera Identification - Do we have a gold standard?},''
\newblock {\em Forensic Science International: Digital Investigation}, vol. 52, pp. 301858, 2025.

\bibitem{BUTORA_2024}
Jan Butora and Patrick Bas,
\newblock ``{The Adobe Hidden Feature and its Impact on Sensor Attribution},''
\newblock in {\em Proceedings of the 2024 ACM Workshop on Information Hiding and Multimedia Security}, New York, NY, USA, 2024, p. 143–148, Association for Computing Machinery.

\bibitem{BARACCHI_2021}
Daniele Baracchi, Massimo Iuliani, Andrea~G. Nencini, and Alessandro Piva,
\newblock ``{Facing Image Source Attribution on iPhone X},''
\newblock in {\em Digital Forensics and Watermarking}, Xianfeng Zhao, Yun-Qing Shi, Alessandro Piva, and Hyoung~Joong Kim, Eds., Cham, 2021, pp. 196--207, Springer International Publishing.

\bibitem{VAZQUEZPADIN_2025}
David Vázquez-Padín, Fernando Pérez-González, and Pablo Pérez-Miguélez,
\newblock ``Apple's synthetic defocus noise pattern: Characterization and forensic applications,'' 2025.

\bibitem{ALBISANI_2021}
C.~Albisani, M.~Iuliani, and A.~Piva,
\newblock ``{Checking PRNU Usability on Modern Devices},''
\newblock in {\em ICASSP 2021 - 2021 IEEE International Conference on Acoustics, Speech and Signal Processing (ICASSP)}, 2021, pp. 2535--2539.

\bibitem{BHAT_2022}
Nabeel Nisar~Bhat and Tiziano Bianchi,
\newblock ``Investigating inconsistencies in prnu-based camera identification,''
\newblock in {\em 2022 IEEE International Conference on Image Processing (ICIP)}, 2022, pp. 851--855.

\bibitem{LUKAS_2005}
Jan Luk{\'a}š, Jessica Fridrich, and Miroslav Goljan,
\newblock ``{Determining digital image origin using sensor imperfections},''
\newblock in {\em Image and Video Communications and Processing 2005}, Amir Said and John~G. Apostolopoulos, Eds. International Society for Optics and Photonics, 2005, vol. 5685, pp. 249 -- 260, SPIE.

\bibitem{CHEN_2008}
Mo~Chen, Jessica Fridrich, Miroslav Goljan, and Jan Luk{\'a}š,
\newblock ``{Determining Image Origin and Integrity Using Sensor Noise},''
\newblock {\em IEEE Transactions on Information Forensics and Security}, vol. 3, no. 1, pp. 74--90, 2008.

\bibitem{MIHCAK_1999}
M.~Kivanc~Mihcak, I.~Kozintsev, and K.~Ramchandran,
\newblock ``Spatially adaptive statistical modeling of wavelet image coefficients and its application to denoising,''
\newblock in {\em 1999 IEEE International Conference on Acoustics, Speech, and Signal Processing. Proceedings. ICASSP99 (Cat. No.99CH36258)}, 1999, vol.~6, pp. 3253--3256 vol.6.

\bibitem{GOLJAN_2009}
Miroslav Goljan, Jessica Fridrich, and Tom{\'a}š Filler,
\newblock ``{Large scale test of sensor fingerprint camera identification},''
\newblock in {\em Media Forensics and Security}, Edward J.~Delp III, Jana Dittmann, Nasir~D. Memon, and Ping~Wah Wong, Eds. International Society for Optics and Photonics, 2009, vol. 7254, p. 72540I, SPIE.

\bibitem{MONTIBELLER_2024b}
Andrea Montibeller, Roy~Alia Asiku, Fernando P\'{e}rez~Gonz\'{a}lez, and Giulia Boato,
\newblock ``{Shedding Light on some Leaks in PRNU-based Source Attribution},''
\newblock in {\em Proceedings of the 2024 ACM Workshop on Information Hiding and Multimedia Security}, New York, NY, USA, 2024, IH\&MMSec'24, pp. 137--142, Association for Computing Machinery.

\bibitem{GSMARENA}
``{GSMArena},'' \url{https://www.gsmarena.com}.

\bibitem{FLOREVIEW_2023}
Daniele Baracchi, Dasara Shullani, Massimo Iuliani, and Alessandro Piva,
\newblock ``{FloreView: An Image and Video Dataset for Forensic Analysis},''
\newblock {\em IEEE Access}, vol. 11, pp. 109267--109282, 2023.

\bibitem{FLICKR_API}
``{Flickr API},'' \url{https://www.flickr.com/services/api}.

\bibitem{TECHREP}
D.~V\'azquez-Pad\'in, F.~P\'erez-Gonz\'alez, and A.~Mart\'in-Del-R\'io,
\newblock ``{{Technical Report: Expanded Analysis of Diagonal Artifacts in Samsung Images}},''
\newblock 2025,
\newblock \url{https://bit.ly/technical_report_ICASSP2026}.

\bibitem{DCRAW}
``{dcraw},'' \url{https://dechifro.org/dcraw}.

\end{thebibliography}

\end{document}